\documentclass[runningheads]{llncs}

\usepackage{eccv}

\usepackage{eccvabbrv}
\usepackage{graphicx}
\usepackage{booktabs}
\usepackage{multirow} 
\usepackage{arydshln} 
\usepackage{svg}
\usepackage{booktabs} 
\usepackage{wrapfig} 
\usepackage{array} 
\usepackage{hyperref}
\usepackage{tabularx}
\usepackage[accsupp]{axessibility}  
\usepackage{orcidlink}

\usepackage{hyperref}

\begin{document}

\title{COSY: Compositional 3DGS Synthesis for Disentangled Human Head Editing} 
\titlerunning{COSY}

\author{
Florian Barthel\inst{1,2}\orcidlink{0009-0004-7264-1672} \and
Shalini De Mello\inst{3}\orcidlink{0009-0009-0213-2860} \and
Koki Nagano\inst{3}\orcidlink{0000-0001-6815-4864} \and
Wieland Morgenstern\inst{2}\orcidlink{0000-0001-5817-7464} \and
Anna Hilsmann\inst{1}\orcidlink{0000-0002-2086-0951} \and
Peter Eisert\inst{1}\orcidlink{0000-0001-8378-4805}
}

\authorrunning{F.\ Barthel et al.}

\institute{Fraunhofer Heinrich-Hertz Institute, Germany \and
Humboldt University Berlin, Germany \and
NVIDIA, USA
}

\maketitle
\begin{figure*}
    \centering
    \includegraphics[width=1.0\textwidth]{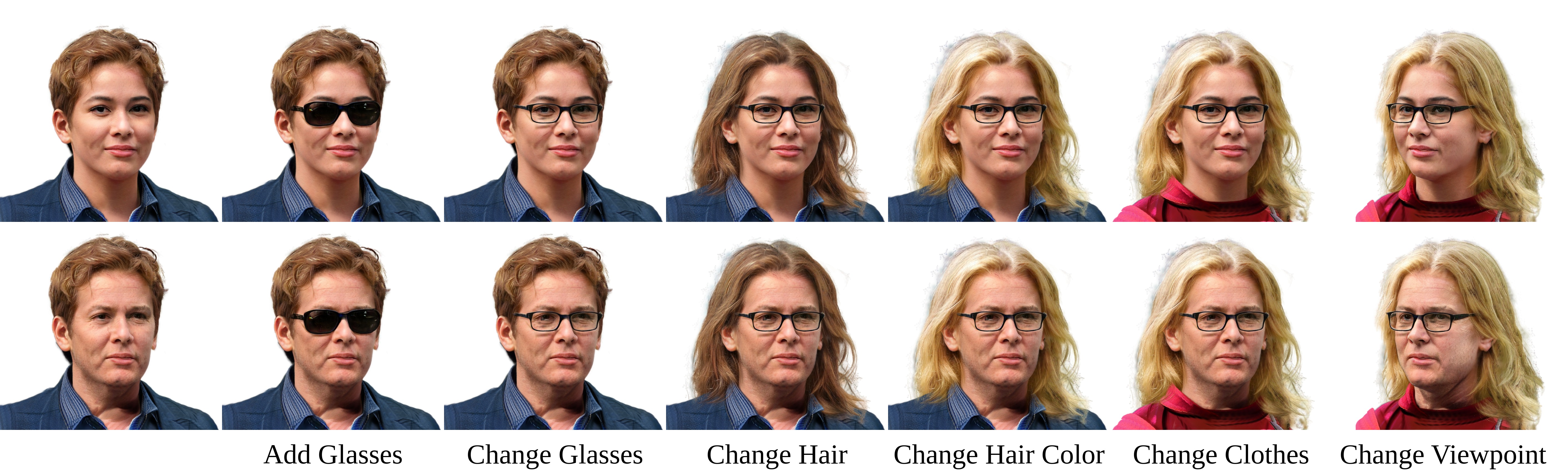}
    \caption{Our method provides an independent latent space for hair, face, glasses, and torso, allowing precise and disentangled editing of 3D human head avatars (\href{https://fraunhoferhhi.github.io/cosy}{Project}).
    }
    \label{fig:teaser}
\end{figure*}

\begin{abstract}
Recent 3D Gaussian Splatting (3DGS) GANs for human heads synthesize and render photorealistic 3D models in real-time and offer a vast variety in identity and appearance. However, controlling specific semantic attributes such as hair color or glasses remains challenging, as edits in the entangled latent space often induce unintended changes in identity or appearance. 
Although there are several methods that aim to disentangle the latent space post training by estimating directions that only modify certain features, these methods cannot guarantee complete disentanglement and often require pre-trained classifiers. 
In our approach, we propose a new generator architecture that synthesizes components, such as hair, skin, glasses, and torso, completely independently.
This allows for changing the latent vector for one region while keeping the remaining parts fixed. Further, we achieve this separation using only sparse information such as the hair or skin color, eliminating the requirement of segmentation masks or geometric priors, often seen in prior work. 
To ensure matching shape and lighting conditions during editing, we allow minimal shared information via context tokens between the independent generators. These tokens even allow us to control the shape and light, without any prior annotation.
Compared to existing works on GAN-based generation and editing, our method shows better disentanglement, more precise editing control, and competitive visual quality. 

\end{abstract}
\section{Introduction}
\label{sec:intro}

The synthesis of realistic 3D human heads is fundamental to film, games, virtual reality, robotics, and interactive content creation. These applications require controllable and disentangled model generation, enabling the user to explore and edit the appearance such as hairstyle or accessories in an interactive manner while preserving the subject’s identity and 3D structure. Controlled 3D synthesis also benefits digital twin pipelines and synthetic data generation (SDG), which depend on scalable methods for producing diverse image, 3D assets and label pairs to support downstream learning and simulation tasks.

Creating and editing high-quality 3D avatar heads, however, often requires extensive manual effort or expensive capturing studios. For this reason, there is a high demand for automating the process while keeping costs to a minimum. 3D GANs have emerged as a suitable framework for this task, as they learn explicit 3D representations exclusively from 2D training data. This makes them stand out from recent diffusion models, which primarily target 2D and video generation or require 3D ground truth data as supervision. Additionally, 3D GANs achieve real time generation capabilities and provide a smooth latent space, that facilitates editing the appearance, also via GAN inversion, which estimates a latent vector that re-synthesizes a specific identity from a given target image. This allows to reconstruct and edit the 3D representation of an identity from a single 2D image. 

Although the rendering quality of 3D GANs has improved substantially with recent advances in NeRF \cite{Chan2021,an2023panohead,xiang2023gram,trevithick2024you} and 3DGS \cite{yu25gaia,kirschstein2024gghead,gsgandecoder,hyun2024gsgan,barthel2025cgsgan,li2025egg3d} based GANs, it still remains challenging to exclusively edit specific attributes of the generated subjects without unintentionally changing other aspects. The core difficulty lies in the unconditional latent space of the generator, in which all semantics are strongly entangled. Although the generator implicitly organizes the latent space during training, enabling editing control through the identification of latent directions corresponding to image attributes \cite{härkönen2020ganspace}, we still observe strong entanglement between the attributes. Another reason for this strong entanglement are biases in the limited training data. For example, in real-world datasets, hair and skin color are highly correlated, making it very difficult to separate these features during editing.

To address these limitations, we propose a new compositional generator architecture that explicitly separates 3D head synthesis into semantically localized components. Our approach uses multiple sub-generators, each responsible for the generation of a specific component, i.e. hair, skin, glasses, and torso. This enables independent control of semantic attributes by modifying the latent vector for a single sub-generator while keeping all others fixed. The separation of components is completely implicit 
without relying on explicit segmentation masks or geometric priors. Further, as we randomly mix components already during the training, we enforce the generator to produce compatible components. In order to encode global attributes such as overall head size and lighting direction, we provide all generators with shared learned features, which even allow us to control the shape and lighting without using any prior annotation. 

Even though training multiple coordinated sub-generators is more challenging compared to generators with a single output, our approach achieves competitive visual quality, while ensuring structurally disentangled editing of head components. Compared to prior work on real-time GAN editing, our work is the first to disentangle the 3D head components using implicit annotations, thus allowing for precise and real-time editing at high resolution. The contributions of our work are:

\begin{enumerate}
    \item A 3DGS GAN architecture that decomposes the synthesis into semantically localized sub-generators, without explicit annotation.
    \item Shared shape and lighting context latent vectors ensuring consistent head geometry and global illumination.
    \item Training strategies to improve the quality of novel component combinations.
\end{enumerate}

\section{Related Work}
\label{sec:related_work}

Generative Adversarial Networks (GANs) \cite{goodfellow2014generative} perform a non-cooperative game between a generator that synthesizes images from random latent vectors and a discriminator that aims to differentiate between generated images and real images from a dataset. During training, both networks are optimized until the generated images look indistinguishable from real images. To better control the synthesized output, conditional GANs \cite{condGAN} allow to include annotations to the training process. Typically, the generator uses the input label to condition the latent vector in order to produce the respective output. To provide conditional feedback, the same label is forwarded to the discriminator, which then decides whether the image is real or fake given the input label.

\subsubsection{3D-aware GANs}
3D-aware GANs are a variant of conditional GANs that use camera parameters as conditioning. However, instead of encoding the latent vector with the camera during generation, the camera parameters are applied to render an intermediate 3D representation produced by the generator. This allows 3D-aware GANs to learn 3D scenes from in-the-wild single-view images. One of the first 3D-aware GANs was $\pi$-GAN \cite{chanmonteiro2020pi-GAN}, which generates the weights of a NeRF \cite{mildenhall2020nerf} MLP. While producing good results, rendering with NeRFs is very slow. As a result, EG3D \cite{Chan2021} and GRAM \cite{xiang2023gram} proposed more efficient 3D representations that store 3D features in a tri-plane (EG3D) or on multiple isosurfaces (GRAM). This allows for much faster synthesis, compared to $\pi$-GAN, while at the same time improving visual quality. Further extensions of EG3D are PanoHead \cite{an2023panohead}, which synthesizes 360° heads using additional training data, or WYSIWYG \cite{trevithick2024you}, allowing for high resolution rendering without super resolution.  

\begin{figure*}[t]
    \centering
    \includegraphics[width=1.0\textwidth]{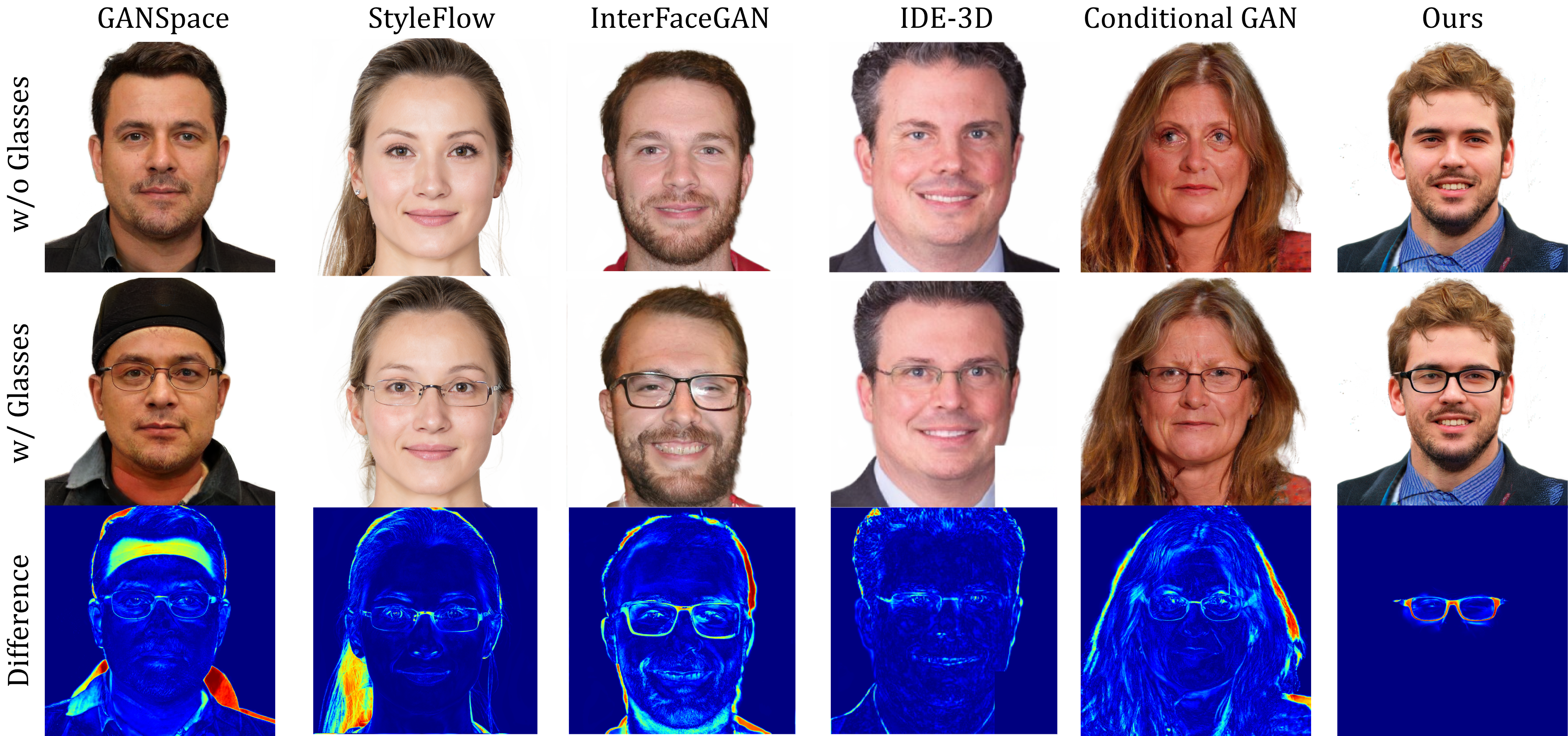}
    \caption{Comparison between GAN editing methods for adding glasses. Existing methods exhibit slight changes, whereas our method cannot change the appearance as the glasses generation is handled in an independent generator branch.}
    \label{fig:glasses_compare}
\end{figure*}

\subsubsection{3DGS GANs}
3D Gaussian Splatting \cite{kerbl3Dgaussians} represents a 3D scene with a set of 3D Gaussian primitives, where every primitive is described by its position, color, covariance, and opacity. Leveraging differentiable rendering, 3DGS optimizes the attributes of all Gaussian primitives to improve over a loss function. This loss can be an image loss when reconstructing a 3D scene from multiple input views, or an adversarial loss when applying 3DGS in a 3D GAN framework. In contrast to NeRFs \cite{mildenhall2020nerf}, 3DGS scenes are explicit and provide very efficient rendering. This allows for real-time rendering and simple integration into VR environments, video games, or even mobile applications. With all these advantages, multiple works have proposed 3DGS based GANs that replace the implicit NeRF representation with an explicit 3DGS scene \cite{gsgandecoder,hyun2024gsgan,li2025egg3d,yu25gaia,li2025egg3d,barthel2025cgsgan} .

A key distinction among the existing 3DGS GAN methods is the initialization of the position. On the one hand, GGHead \cite{kirschstein2024gghead}, GAIA \cite{yu25gaia}, and EGG3D \cite{li2025egg3d} synthesize UV maps with attributes and position offsets of Gaussians on a template mesh derived from FLAME \cite{FLAME}. This template-based formulation produces high quality visual results and even enables animation control with GAIA \cite{yu25gaia} and EGG3D \cite{li2025egg3d}. On the other hand, approaches like GSGAN \cite{hyun2024gsgan} and CGS-GAN \cite{barthel2025cgsgan} model Gaussian positions as an unconstrained point cloud, optimized by a transformer architecture inspired by point-cloud generative models \cite{guo2021point,zhao2021point,nichol2022point}. To impose structural coherence within the unconstrained point cloud representation, both GSGAN and CGS-GAN leverage a hierarchical structure of Gaussian primitives in which smaller Gaussian primitives are positioned relative to larger ones. As a result, both methods produce state-of-the-art results without relying on a predefined head template. Further, CGS-GAN \cite{barthel2025cgsgan} eliminates explicit view conditioning, which could lead to representations that appeared realistic primarily from specific viewing directions in previous 3DGS GANs. By eliminating this dependency, CGS-GAN encourages the generator to learn a view-consistent 3D structure, thereby improving multi-view coherence. This is demonstrated by the $\text{FID}_\text{3D}$ metric that evaluates the 3D generator without informing the model in advance from which viewing direction the 3D model will be rendered.

Our approach builds upon CGS-GAN, as it achieves strong multi-view 3D consistency among existing unconstrained 3DGS GANs and does not rely on predefined template meshes. This is particularly important, since we do not assume the availability of template geometries for hair, glasses, face, or torso.

\subsubsection{Latent Space Editing}
Editing is often described as a core strength of GANs compared to multi-view reconstruction methods or diffusion models. However, achieving strictly disentangled editing control over individual attributes remains challenging. As GANs are trained to approximate the full data distribution, the unconditional latent space captures dataset specific correlations and biases. For human head generation, the latent space strongly entangles hair and skin color, making it difficult to change one without the other. To counteract this, several methods aim to separate the latent space after training into regions that correlate with interpretable image features. GANSpace \cite{härkönen2020ganspace} computes PCA components of the latent space and heuristically assigns the components to face features. Typically, the first components affect major modifications, such as the identity, while minor components adjust subtle features, like lighting or expression. Although this strategy enables intuitive edits without external guidance, it provides only limited control and strong attribute entanglement.
InterFaceGAN \cite{shen2020interfacegan} extends the idea by identifying  attribute-specific editing directions in the latent space using a pre-trained classifier. Similarly, StyleFlow \cite{abdal2021styleflow} uses a pre-trained classifier to remap the entangled latent space to a learned conditional latent space. Although classifier-based editing methods generally improve attribute controllability, they remain constrained by the underlying entangled representation. 
An example of this strong entanglement is shown in Fig. \ref{fig:glasses_compare}. Here, we observe slight changes in identity or hairstyle when adding glasses.

\subsubsection{Reference based Editing}
Reference based editing methods aim to combine attributes of two input images. For example, Barbershop \cite{zhu2021barbershop} and HairCLIPv2 \cite{wei2023hairclipv2} transfer the hairstyle of a reference person onto a target identity. Similarly, StyleFusion \cite{kafri2021stylefusion} or TriplaneEdit \cite{bilecen2025reference} fuse features of two inputs and enable precise feature swapping guided by pre-trained masking networks. These methods achieve improved disentanglement compared to latent space editing, however, do not allow free exploration of the latent space, given the requirement of a reference image as input. Further, unlike latent space editing methods that operate in real-time, these approaches often take several minutes \cite{zhu2021barbershop} to compute a single output image.

\subsubsection{Compositional Generation}
The idea of decomposing the generation into multiple generators has been explored in prior works. For example, GIRAFFE \cite{xue2022giraffe}, GSGAN \cite{hyun2024gsgan} and EGG3D \cite{li2025egg3d} split the generation of foreground and background using separate generators. In addition, separating hair and face generation was achieved in 3DGH \cite{he20253DGH} and EGG3D \cite{li2025egg3d}. In contrast to our approach, however, existing methods either constrain the separation with explicit geometric priors, such as positioning skin Gaussians on the surface of a FLAME head \cite{li2025egg3d}, or using a segmentation mask during training to guide the separation \cite{he20253DGH}.

Similarly to compositional generation, several approaches \cite{chen2022sofgan,sun2022ide,yun2025ffacenerf,sun2022fenerf,zhu2020sean} learn a direct mapping between 2D segmentation masks and generated images. This allows drawing or editing the outline of face regions, like hair, nose, or mouth, which then results in a realistic representation with same proportions.

Among existing GAN editing methods, our approach is mostly related to latent editing methods.
Instead of shaping the proportions with segmentation masks or fusing attributes of reference images, our method splits the generation process semantically, allowing full exploration of the latent space of each component in real-time. For this reason, we mainly compare the editing quality with latent space methods and conventional conditional GANs.

\subsubsection{Distinction to Diffusion Models}
Recent diffusion models \cite{labs2025flux1kontextflowmatching,gong2025seedream} for image and video synthesis have surpassed the quality of 2D GANs \cite{Karras2019stylegan2,brock2018large,sauer2022stylegan}. Nevertheless, in the context of 3D generation, GANs remain particularly attractive, as they can learn 3D representations directly from 2D image collections without 3D ground truth. Diffusion models, in contrast, inherently require a 3D ground truth for the noising and denoising step during training. 
Although several works have applied diffusion models in 3D head avatar generation \cite{lan2023gaussian3diff,chen2024morphable}, they either train on synthetic 3D data (generated by 3D GANs \cite{lan2023gaussian3diff}), or use an intermediate model \cite{liu2023zero,liu2023syncdreamer} to convert 2D inputs into 3D pseudo ground truth. Further, GANs offer a smoother latent space \cite{zhang2024diffmorpher} and allow for real-time synthesis.

\begin{figure*}[t]
    \includegraphics[width=1.0\textwidth]{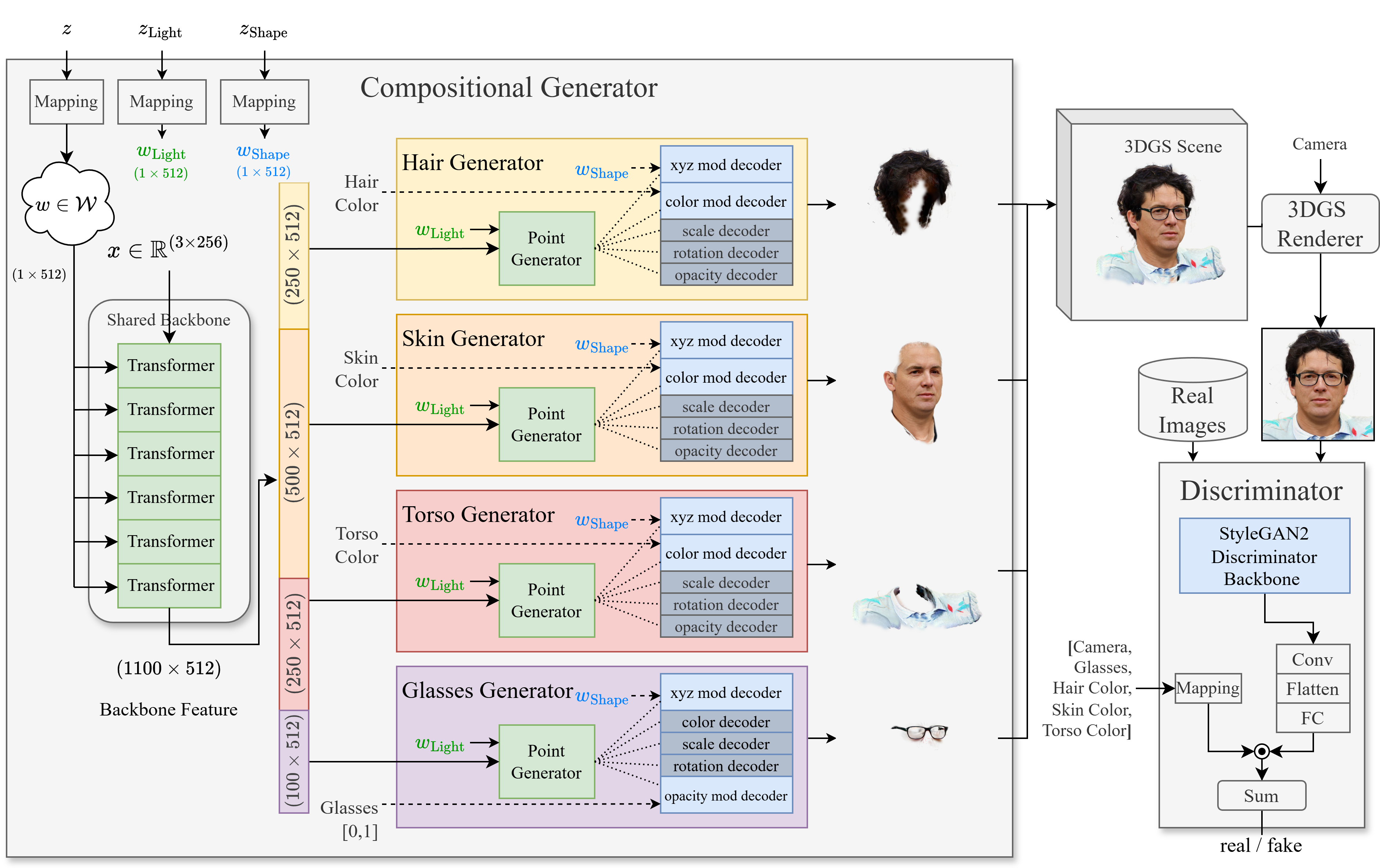}
    \caption{Overview of our compositional 3D GAN architecture. Initially, we generate shared features with a transformer backbone that we partition and distribute to the sub-generators. Each sub-generator then produces a 3DGS scene using an adapted CGS-GAN \cite{barthel2025cgsgan} generator. In the final decoder layer, we modulate the colors given the conditioning from the dataset. Additionally we provide limited shared information $w_\text{Shape}$ and $w_\text{Light}$ to allow coordination along shape and lighting.}
    \label{fig:arch}
\end{figure*}

\section{Method}
To address the limitations of latent editing methods and entangled conditional GANs, we propose a compositional 3D GAN architecture that enables component-level control, using sparse conditions like colors or boolean flags. Our method consists of two main ideas. Firstly, a compositional GAN architecture to produce disentangled 3D components. Secondly, several techniques to improve the quality and consistency when mixing different components during inference.

\subsection{Compositional GAN Architecture}
\label{sec:arch}
To enable independent manipulation of semantic attributes, we construct the generator as a composition of multiple sub-generators. As shown in Fig. \ref{fig:arch}, we allocate a dedicated 3DGS generator, similar to CGS-GAN \cite{barthel2025cgsgan}, to each semantic region, i.e., hair, face, glasses and torso. The complete 3D head is then formed by composing the Gaussian primitives into a single coherent scene representation. The composed 3D scene is then rendered into a 2D image using a differentiable 3DGS renderer, which is forwarded to the discriminator. To enforce component-wise specialization, we supply region-specific attribute annotations only to the corresponding sub-generator. Here, we differentiate between additive components like glasses, which are turned on and off, and permanent components like the skin, which are present in all head models.

\subsubsection{Additive Components}
To describe components, which are not always present during generation (like glasses), we use a boolean flag that describes the presence or absence of this component. If the component should be present, we generate it and add it to the composition, while we hide it by subtracting a large value from the Gaussian opacity when it is absent. At the same time, we provide the boolean flag as 0 or 1 to the discriminator, which penalizes the generator if the respective component is incorrectly missing or appearing. As the boolean flag is exclusively provided to the respective sub-generator, it implicitly learns to become responsible for generating this component. Similarly, if the sub-generator starts to generate more than necessary, for example, if the glasses generator produces the eyes or the skin, those features will be missing from the composition, when the output is hidden. As a result, the generator learns to precisely limit the generation to the designated component.

\begin{figure*}[t]
    \centering
    \includegraphics[width=1.0\textwidth]{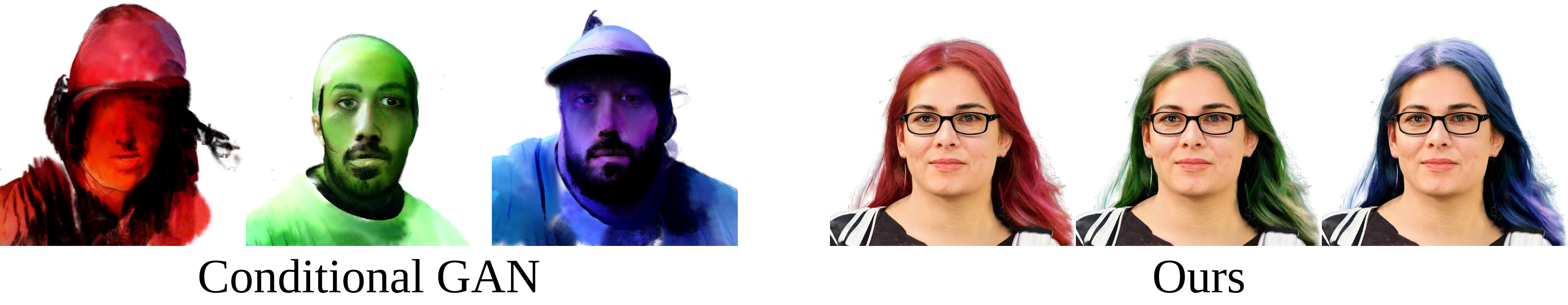}
    \caption{Failure example of conditional GANs when providing colors that have never been seen in training. With conditional GANs, such inputs push the latent code into unexplored regions resulting in drastic image changes. Our method directly converts the input into Gaussian color attributes, leaving the geometry unchanged.}
    \label{fig:novel_colors}
\end{figure*}

\subsubsection{Permanent Components}
To synthesize components always present during generation (like the skin), we have to define a description that exclusively relates to the respective region. 
For our purpose, we select the color as a suitable description to separate skin, hair, and the torso. Additionally, this gives us the ability to control the color next to the overall latent vector. Specifically, we use a sparse 10-bin RGB histogram aggregated over the respective head region. This color histogram is then applied to modulate the color attribute of the Gaussian primitives inside the respective sub-generator. This is done in the \textit{color mod decoder} layer, shown in Figure \ref{fig:arch}, which converts the features of the point generator into Gaussian color attributes. Similar to the additive component, we provide the same color information to the conditional discriminator, which then produces a loss if the target color is not met. This sparse information is sufficient to cleanly separate the components of our 3D human head into hair, skin and torso. By only modulating 3DGS color attributes, we achieve very good generalization for novel colors. As shown in Figure \ref{fig:novel_colors}, our generator learns a direct mapping from color to 3DGS features, whereas naive conditional GANs that condition the latent vector on an input conditioning usually break when provided with out-of-distribution conditions.

\subsubsection{Discriminator}
We employ the conditional discriminator architecture of EG3D \cite{Chan2021} with minimal changes. In addition to the camera conditioning, we concatenate all color information and the boolean glasses flag into one single vector and multiply it to the discriminator features.
In contrast to existing compositional GANs \cite{li2025egg3d,he20253DGH}, our discriminator receives the entire image instead of a masked region, as the semantic separation is entirely handled within the generator. We consider this design advantageous, as segmentation masks are often not precise, while additionally mask-based supervision is difficult to apply during early training stages where the generator produces less structured outputs.

\subsection{Global Consistency}
With the architecture described in \ref{sec:arch}, the generator is now able to separately synthesize hair, glasses, face, and torso. However, with this design, the components will not fit well together, when synthesizing novel combinations with different latents in each sub-generator during inference. To enforce better compatibility, we propose several techniques that improve global consistency in shape and light, while also preparing the generator for novel component combinations already during training.

\subsubsection{Latent Mixing}
\label{sec:mixing}
The most straightforward way to enforce better compatibility during inference is by already producing mixed components during training. This is achieved by randomly swapping the backbone features of one component during the forward pass in the generator. This process essentially simulates the editing of a generated 3D head during inference. We then combine the mixed 3DGS scenes into one scene by concatenating the 3DGS features and forward the renderings to the discriminator. The discriminator then produces a training signal that penalizes the generator if the components do not fit together. This avoids generating gaps or overlaps between mixed components.

\begin{figure*}[t]
    \centering
    \includegraphics[width=1.0\textwidth]{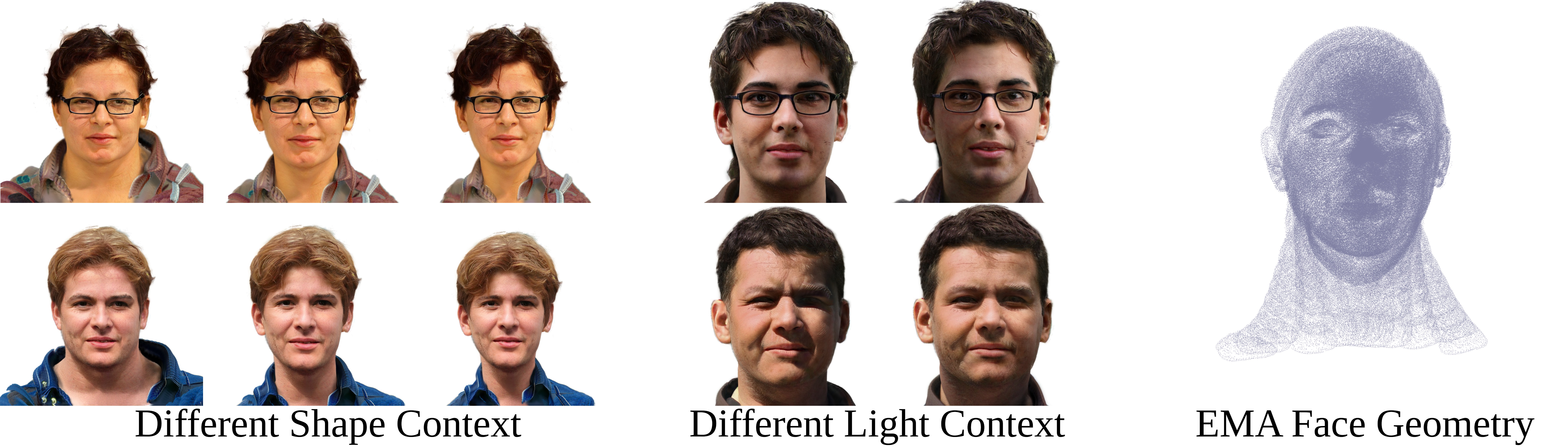}
    \caption{Demonstration of different shape context vectors (left), light context vectors (middle) and the EMA face geometry (right).}
    \label{fig:context_token}
\end{figure*}

\subsubsection{Shape Context}
With the latent mixing method introduced in \ref{sec:mixing}, we enforce better compatibility between components that were synthesized from different latent vectors. This, however, poses the challenge that in the training dataset, not all hairstyles, glasses or torso can possibly fit together. For example, the hair of a young child with a small head will likely not fit on a very large head. To avoid that the generator collapses to a single head shape that always fits together, we introduce a small latent code ($w_\text{Shape} \in \mathbb{R}^{512}$ in Fig. \ref{fig:arch}) that we share across all sub-generators and is never mixed during latent mixing. To make sure that the sub-generators do not use this shared signal to communicate anything but the shape, we forward this latent code only to the final xyz decoder layer in each sub-generator. This way, all other 3DGS attributes cannot be changed. Further, as the final position decoder is constructed as a small MLP with limited capacity, the model is prevented from encoding complex identity changes through this latent code. Fig. \ref{fig:context_token} demonstrates the effect of only changing the shape context vector, resulting in different head shapes but consistent identity.

\subsubsection{Light Context}
Similarly to inconsistencies in shapes and sizes, all sub-generators have to agree on a shared global lighting condition. 
Otherwise, each sub-generator may synthesize different directional lighting, resulting in visually implausible compositions. 
Therefore, we provide the sub-generators with a shared low dimensional lighting latent code (see $w_\text{Light} \in \mathbb{R}^{512}$ in Fig. \ref{fig:arch}) that is never mixed during latent mixing. Intuitively, to achieve this, we would simply modulate the output colors in the decoder layers, similarly to the shape context vector. However, as the final color decoder layer provides very limited capacity, it cannot model complex shadows or reflections. For this reason, we condition the input latent vector with the light context vector. As this allows the generator to modify everything, and not just the lighting, we then apply a regularization that penalizes the generator if two renderings from the same latent, but with different light context, differ too much. Here, we use the LPIPS distance of two images, rendered from the same camera pose. If the regularization strength is set correctly, we find that the generator implicitly learns to use this shared vector exclusively for lighting changes, as demonstrated in Fig. \ref{fig:context_token}.

\subsubsection{Exponential Moving Average Regularization}
In a compositional generator setting, one component may under-represent regions that are statistically also occupied by another component. In our case, the face generator may reduce coverage near the hairline, implicitly relying on the hair generator to complete the geometry. This leads to geometric inconsistencies when novel component combinations are synthesized. To avoid this, we introduce a geometric consistency loss penalizing deviations of the generated face geometry from an exponential moving average (EMA) face geometry. Specifically, we track the \textit{xyz} coordinates and the opacity values, which are always in the same order. As the EMA face captures the entire forehead (Fig. \ref{fig:context_token} right), given that the majority of faces have a visible forehead, this loss pushes the generator towards synthesizing a more complete face.

\section{Experiments}
We compare our results from two main points of view. First, we evaluate the overall visual quality, compared to recent unconditional 3DGS GANs, and second, we compare the editing performance to existing GAN editing methods. Specifically, we measure the disentanglement, while also evaluating how well the output image represents the target label. Finally, we apply several ablations and perform a latent space analysis with GANSpace \cite{härkönen2020ganspace} on the sub-latent spaces of each component. Further examples and training details are provided in the supplementary materials.

To compare the visual quality and editing performance with a conventional conditional GAN \cite{condGAN,Karras2019stylegan2}, we provide an additional baseline, in which we train the most recent 3DGS GAN (CGS-GAN \cite{barthel2025cgsgan}) with the same color and glasses conditioning. Here, however, the conditioning is directly forwarded as a single vector to the generator mapping network, while using the same discriminator.

\subsubsection{Dataset}
We use FFHQC \cite{barthel2025cgsgan}, which is a subset of FFHQ \cite{Karras2019stylegan2} with less occluding objects and larger cropping. 
To extract color information for the hair, skin and toros,  we use BiSeNet \cite{yu2018bisenet} to extract rough regions and compute a 10-bin histogram on each color channel. We then concatenate all three histograms into one 30-dimensional vector that serves as the color conditioning for our compositional training. As we do not need precise segmentation masks, we shrink the respective regions, to avoid leaking color information to adjacent regions. Finally, we obtain the binary glasses flag from the FFHQ-features repository \cite{Dcgm}.

\subsubsection{Metrics}
We use the Fréchet Inception Distance (FID) \cite{heusel2017gans} to estimate the visual quality of the generated images.
FID computes the difference between Inception-V3 \cite{inceptionv3} features for synthesized and real images. A low value indicates a very similar distribution, suggesting good visual quality. In particular, to measure the quality of the 3D heads, we employ the $\text{FID}_{\text{3D}}$ metric, which evaluates FID without using camera view conditioning to a generator. Essentially, $\text{FID}_{\text{3D}}$ hides the camera pose from the generator during the synthesis, which better reflects the actual 3D quality. 

To evaluate editing disentanglement, we measure the extent to which editing one attribute affects unrelated regions. For this evaluation, we use the face parsing network BiSeNet \cite{yu2018bisenet} to obtain the regions that should not change during editing. On those patches, we then compute the LPIPS distance before and after editing. A low value denotes low attribute entanglement.
In addition, we compute a recall for how often we see glasses when the glasses label was activated.

\begin{table}[b]
    \caption{Unconditional image quality measured with $\text{FID}_{\text{3D}}$ across recent 3DGS GAN methods using the FFQHC dataset at 512 and 1024 resolution.}
    \centering
    \begin{tabular}{l >{\raggedleft\arraybackslash}p{2cm} >{\raggedleft\arraybackslash}p{2cm}}
        \toprule
        $\textbf{FID}_{\textbf{3D}}$ & $\downarrow$ $512^2$ & $\downarrow$ $1024^2$\\
        \midrule
        GSGAN \cite{hyun2024gsgan} & 7.68 & - \\
        GGHead \cite{kirschstein2024gghead} & 7.78 & 14.27  \\
        CGS-GAN \cite{barthel2025cgsgan} & 4.53 & \textbf{5.25} \\
        Cond CGS-GAN & 4.52 & -  \\
        Ours & \textbf{4.42} & 5.43  \\
        \bottomrule
    \end{tabular}
    \label{tab:fid_table}
\end{table}

\subsubsection{Results on Visual Quality}
\begin{figure*}[t]
    \centering
    \includegraphics[width=1.0\textwidth]{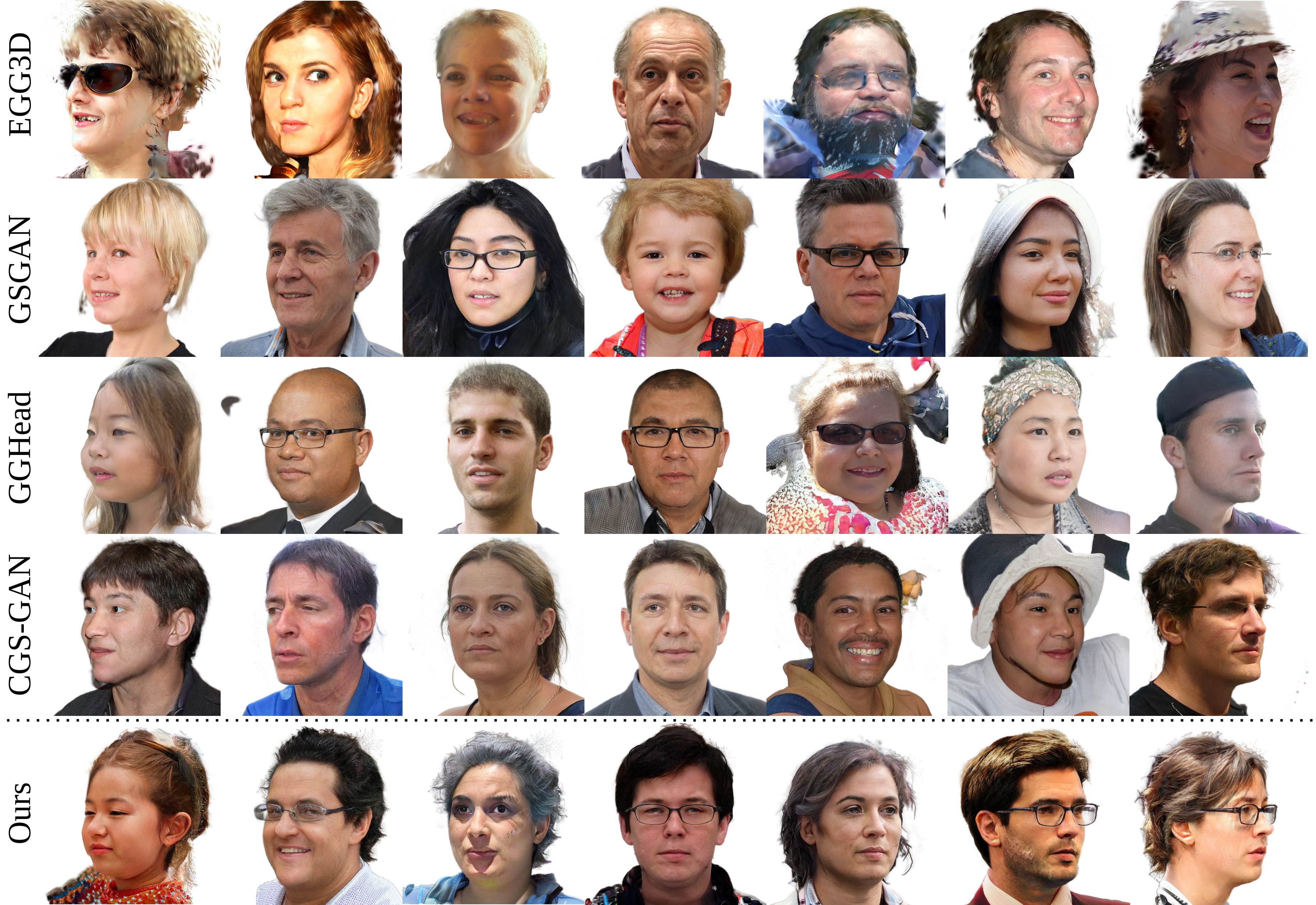}
    \caption{Visual comparison to recent 3DGS GANs at $512^2$ resolution with truncation $\psi=0.8$. EGG3D \cite{li2025egg3d}, GSGAN \cite{hyun2024gsgan} and GGHead \cite{kirschstein2024gghead} produce worse side views, as those methods condition the generation on the viewing direction, which we set to a fixed frontal view.
    For EGG3D \cite{li2025egg3d}, we visualize their results without background using their official checkpoint that is trained with FFHQ with background.}
    \label{fig:vis_quality}
\end{figure*}

In Tab. \ref{tab:fid_table} we compare the $\text{FID}_{\text{3D}}$ metric with existing unconditional 3DGS GANs, as well as our conditional GAN baseline. Although the training is much more challenging, given that our generator has to coordinate the separation across different sub-generators, we achieve state-of-the-art $\text{FID}_{\text{3D}}$ results with 4.42 on $512^2$ resolution and 5.43 on $1024^2$ resolution. As our method synthesizes no backgrounds and FID is sensitive to background~\cite{Kynkaanniemi2022}, we cannot directly compare it to EGG3D \cite{li2025egg3d} (FID 7.51 with FFHQ 512 resolution with background). 
A visual comparison is shown in Fig. \ref{fig:vis_quality}. Like CGS-GAN \cite{barthel2025cgsgan}, our model produces high quality renderings with consistent side view performance. In contrast, EGG3D \cite{li2025egg3d}, GSGAN \cite{hyun2024gsgan} and GGHead \cite{kirschstein2024gghead} produce lower quality towards steep viewing angles, as they condition the 3D synthesis on a specific viewing direction. 

\subsubsection{Results on Editing Quality}
\begin{figure*}[t]
    \centering
    \includegraphics[width=1.0\textwidth]{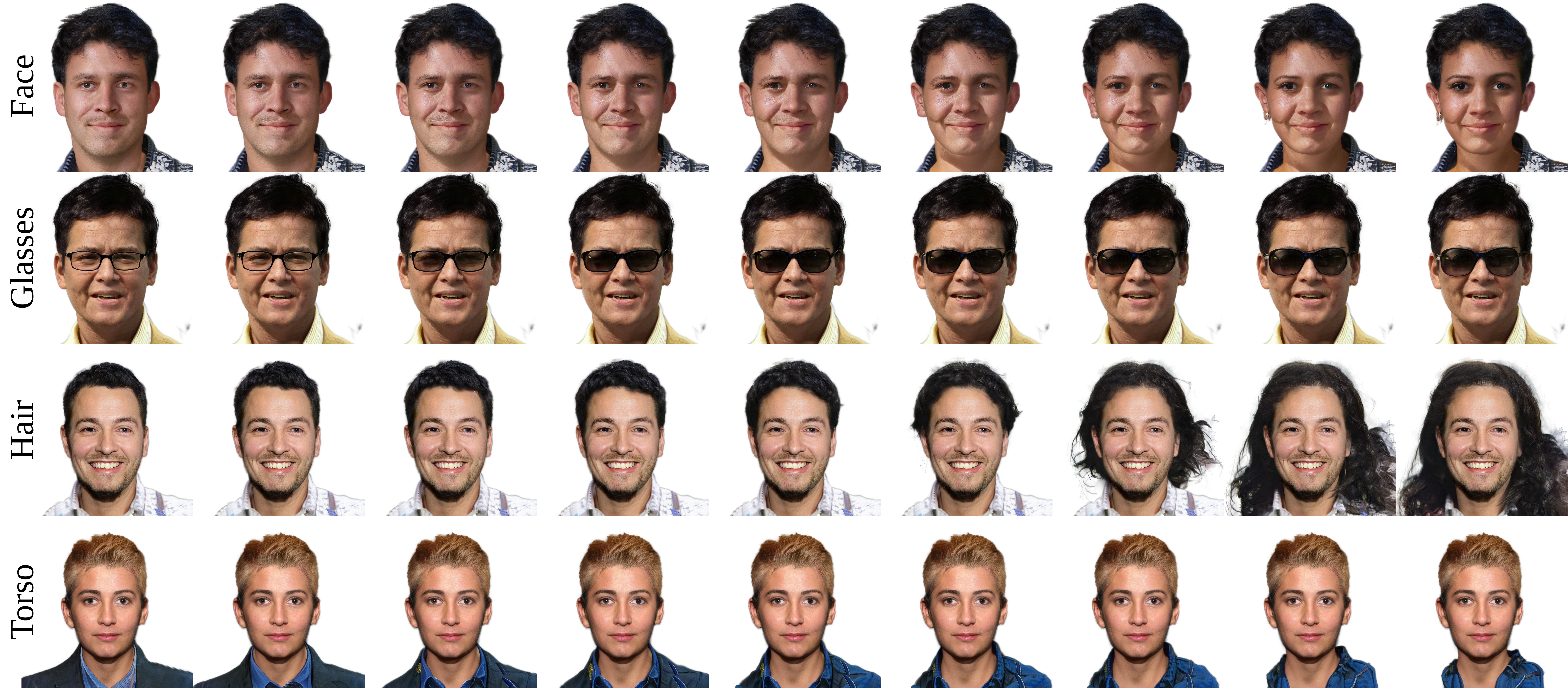}
    \caption{Traversing along the first PCA direction of each generator. Next to overall identity changes in the face, we observe a semantic separation between large sunglasses and small reading glasses, long and short hair, and formal cloth and an open jacket.}
    \label{fig:pca}
\end{figure*}

The strong side of our method lies in the editing performance compared to existing real-time GAN editing methods. As our model generates different components with separate sub-generators, we allow changing one attribute without touching the others. This is demonstrated in Fig. \ref{fig:teaser}, \ref{fig:glasses_compare}, and \ref{fig:pca}, while we also supported our results numerically in Tab. \ref{tab:lpips}. Here, we add glasses or change the hairstyle using the respective editing method and measure the LPIPS distance of before and after editing in unrelated regions. We observe significant perceptual changes, especially with GANSpace \cite{härkönen2020ganspace} or with the conditional GAN baseline. Our approach in contrast produces very little changes. The value is not 0 as the model produces small shadows under the glasses which are measured as slight perceptual differences.

\begin{table}[b]
\parbox{.5\linewidth}{
    \caption{We measure the LPIPS distance in unrelated regions after adding glasses or changing the hairstyle.}
    \centering
    \small
    \begin{tabular}{lcc}
        \toprule
        \textbf{LPIPS Stability} & Glasses $\downarrow$ & Hair $\downarrow$ \\
        \midrule        
        GANSpace        \cite{härkönen2020ganspace} & 0.254 & 0.165 \\
        InterFaceGAN    \cite{shen2020interfacegan} & 0.196 & - \\
        StyleFlow       \cite{abdal2021styleflow}   & 0.172 & 0.133  \\
        Cond CGS-GAN   & 0.239 &  0.149 \\
        Ours   & \textbf{0.016} & \textbf{0.014}  \\
        \bottomrule
    \end{tabular}
    \label{tab:lpips}
}
\hfill
\parbox{.4\linewidth}{
    \caption{Recall on how often glasses are present when the glasses label is active.}
    \centering
    \small
    \begin{tabular}{lc}
        \toprule
        \textbf{Editing Recall} & Glasses $\uparrow$ \\
        \midrule
        GANSpace \cite{härkönen2020ganspace} & 0.508  \\
        InterFaceGAN \cite{shen2020interfacegan} & 0.632  \\
        StyleFlow \cite{abdal2021styleflow} & 0.948 \\
        Cond CGS-GAN &  0.952 \\
        Ours & \textbf{0.997} \\
        \bottomrule
    \end{tabular}
    \label{tab:recall}
}
\end{table}

In Tab. \ref{tab:recall}, we measure how often the editing method actually produces the target label. Specifically we select 1000 images without glasses and apply the respective editing method to add glasses. Here, GANSpace \cite{härkönen2020ganspace} and InterFaceGAN \cite{shen2020interfacegan} show the least reliable results, as they only show glasses in about 50\% or 60\%. Although a higher recall could be gained from those methods, it would come with a tradeoff in LPIPS stability, which we calculate in  Tab. \ref{tab:lpips} with the same image pairs. StyleFlow \cite{abdal2021styleflow} and our conditional GAN baseline show much stronger results by achieving a recall of about 95\%. Our method achieves almost 100\% recall, as the model simply turns on the respective glasses sub-generator when the glasses label becomes active.

\subsubsection{Sub-latent Space Analysis}
To gain a better intuition on how the latent space of each sub-generator behaves, we apply the GANSpace \cite{härkönen2020ganspace} approach on top of our method. Specifically, we compute the PCA components for each latent space and perform linear interpolation along the main directions. The result is demonstrated in Fig. \ref{fig:pca}. Here, we observe that each latent space encodes semantic features along the main PCA components. When changing the face latent code, we observe a shift from masculine to feminine appearance. Further, the glasses latent space separates large sunglasses and small reading glasses along the main PCA component. Similarly, the hair is changed from long hair to short hair, and the torso shifts from an open jacket to a closed suit.

\subsubsection{Light Context Ablation}
The strength of the light context regularization determines how perceptually different we allow the components to be when changing only the shared lighting context vector. Here, our target is to set the strength to an appropriate value so that large changes in hairstyle or identity are eliminated, while subtle changes like light and shadows are allowed. In a series of experiments we find that an optimal tradeoff is found when setting the value in a way that the mean LPIPS distance is about 0.3. This is shown in Fig. \ref{fig:ablation} (left) where we observe too large changes when the regularization weight is set too small, and no changes at all if regularization weight is too high. Although this parameter is very sensitive, it does not affect the overall training stability.

\subsubsection{EMA Geometry Loss Ablation}
\begin{figure*}[t]
    \centering
    \includegraphics[width=1.0\textwidth]{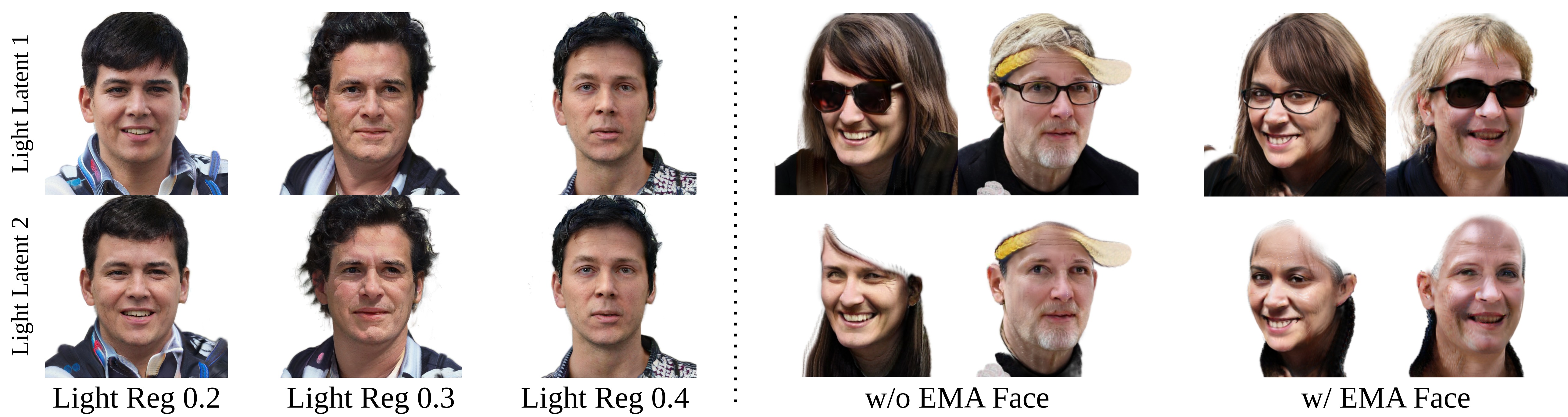}
    \caption{Left: Ablation of different light context regularization strengths, with 0.3 showing a good tradeoff between too much and too little changes. Right: Ablation without the EMA face regularization, which produces incomplete foreheads (the images are taken from an early training checkpoint).}
    \label{fig:ablation}
\end{figure*}

Similarly to the shape context, we encourage the generator to produce matching components by regularizing the distance between the generated face geometry and an EMA face geometry. As demonstrated in Fig. \ref{fig:ablation} (right), this avoids generating incomplete face geometries that stop just below the matching hairline, while also preventing the head generator from synthesizing outliers like headward.

\begin{wraptable}{r}{0.4\textwidth}
    \vspace{-36pt}
    \centering
    \caption{FID and $\text{FID}_\text{Mix}$ results with different latent mixing probabilities.}
    \label{tab:mixing}
    \begin{tabular}{lcc}
        \toprule
        Mixing Prob. & FID $\downarrow$ & $\text{FID}_\text{Mix}$ $\downarrow$\\
        \midrule
        0\%  & 4.88 & 10.51 \\
        10\% (Ours) & \textbf{4.42} & 8.32 \\
        20\% & 4.98 & 9.07 \\
        30\% & 5.16 & 8.49  \\
        40\% & 5.97 & \textbf{6.53}  \\
        \bottomrule
    \end{tabular}
    \vspace{-12pt}
\end{wraptable}

\subsubsection{Latent Mixing Ablation}
By synthesizing mixed examples during training, we ensure better compatibility between components during inference. In Tab. \ref{tab:mixing}, we test different mixing probabilities. A too high mixing probability harms the overall training, resulting in lower FID, while a too little mixing produces poor visual results when editing during inference. To measure this, we calculate another FID score, where we only evaluate the quality of mixed compositions. For our final method we select 10\% as it yields a good tradeoff between FID and $\text{FID}_\text{Mix}$.

\section{Conclusion}

We present a compositional 3D GAN architecture that decomposes the 3D human head generation into semantically localized components. With our design that avoids explicit geometric priors or segmentation masks, the generator implicitly learns a clean separation from colors or boolean flags. 
This allows for precise GAN editing of hairstyle or glasses without modifying the rest. 
By allowing the sub-generators minimal communication using shared context latent codes, we improve the generalization while achieving manual control over lighting and shape without prior annotation. Further, we apply two training regularizations that improve the visual quality when composing a 3D head from different latent codes. In our experiments, we achieve competitive visual quality measured with FID, while achieving low entanglement and high editing accuracy.

In future work, the compositional framework could be extended to additional components, such as hats, beards, or accessories. Furthermore, none of the components in our generator architecture are specifically designed for 3D human heads. Hence, it can be adapted to other domains where objects can be decomposed into meaningful parts. This can include full body avatars with disentangles body and cloth for virtual try on, or 3D cars with customizable parts. Such extensions allow the method to be deployed for synthesizing annotated  3D simulation environments with a wide variety.

\section*{Acknowledgements}
This work has partly been funded by the German Research Foundation (project 3DIL, grant
no.\ 502864329), the European Commission (Horizon Europe project Luminous, grant no.\ 101135724), and the BMFTR - Federal Ministry of Research, Technology
and Space (project REFRAME, grant no. 01IS2407A).

\bibliographystyle{splncs04}
\bibliography{main}
\end{document}